\documentclass[letterpaper, 10 pt, conference]{ieeeconf}  % Comment this line out if you need a4paper

\IEEEoverridecommandlockouts                              % This command is only needed if 
\usepackage{graphicx}      % include this line if your document contains figures
\usepackage{algorithm} % added for algorithm
\usepackage{algpseudocode} % added for algorithm
\usepackage{varwidth} % add for algorithm
\usepackage{cite}
\usepackage{diagbox}
\usepackage{subcaption}
\usepackage{amsfonts}
\usepackage{amsmath}
\usepackage{url}
\usepackage{gensymb}
\usepackage{tabularx,caption}
\usepackage[dvipsnames]{xcolor}
\usepackage{hyperref}
\newcommand{\argmin}{\mathop{\mathrm{argmin}}}  
  
\graphicspath{{figs/}}
\title{\LARGE \bf
Allowing Safe Contact in Robotic Goal-Reaching: \\ Planning and Tracking in Operational and Null Spaces
}

\author{Xinghao Zhu$^{1,2}$, Wenzhao Lian$^3$, Bodi Yuan$^2$, C. Daniel Freeman$^4$, and Masayoshi Tomizuka$^1$
\thanks{$^1$ Mechanical Systems Control Lab, Mechanical Engineering, UC Berkeley, CA, USA
{\tt\small \{zhuxh,tomizuka\}@berkeley.edu}}
\thanks{$^2$ X, The Moonshot Factory, CA, USA
{\tt\small bodiyuan@google.com}}
\thanks{$^3$ Intrinsic Innovation LLC, CA, USA
{\tt\small wenzhaol@google.com}}
\thanks{$^4$ Google Research, CA, USA
{\tt\small cdfreeman@google.com}}
}

\begin{document}
\maketitle
\thispagestyle{empty}
\pagestyle{empty}

\begin{abstract}
In recent years, impressive results have been achieved in robotic manipulation. While many efforts focus on generating collision-free reference signals, few allow safe contact between the robot bodies and the environment. However, in human's daily manipulation, contact between arms and obstacles is prevalent and even necessary. This paper investigates the benefit of allowing safe contact during robotic manipulation and advocates generating and tracking compliance reference signals in both operational and null spaces. In addition, to optimize the collision-allowed trajectories, we present a hybrid solver that integrates sampling- and gradient-based approaches. We evaluate the proposed method on a goal-reaching task in five simulated and real-world environments with different collisional conditions. We show that allowing safe contact improves goal-reaching efficiency and provides feasible solutions in highly collisional scenarios where collision-free constraints cannot be enforced. Moreover, we demonstrate that planning in null space, in addition to operational space, improves trajectory safety. Further information is available at~\href{https://rolandzhu.github.io/ContactReach/}{https://rolandzhu.github.io/ContactReach/}.
\end{abstract}

%%%%%%%%%%%%%%%%%%%%%%%%%%%%%%%%%%%%%%%%%%%%%%%%%%%%%%%%%%%%%%%%%%%%%%%%%%%%%
%%%%%%%%%%%%%%%%%%%%%%%%%%%%% Introduction Section %%%%%%%%%%%%%%%%%%%%%%%%%%
%%%%%%%%%%%%%%%%%%%%%%%%%%%%%%%%%%%%%%%%%%%%%%%%%%%%%%%%%%%%%%%%%%%%%%%%%%%%%

\section{Introduction}
\label{sec: introduction}
Robotic manipulation in unconstrained environments has received increasing attention in recent years.
% Robotic manipulation has been researched and deployed to address variant requirements.
Although much research focuses on generating and tracking collision-free reference signals (e.g., the desired trajectory)~\cite{220713438, 9811973, 9561954, martin2019iros}, some manipulation tasks benefit from allowing contact between the robot and the environment. For example, in shelf picking, the robot sometimes must collide with obstacles to grasp the target object~\cite{5652970, andy_grasp_push}; in agricultural harvesting, it is favourable for the robot to push away the occluding petioles and leaves~\cite{s40648}. 

% \wenzhao{Fine to remove this example as it's not as intuitive as the previous two. In industrial assembly, the workspace might be protected by deformable curtains that require the robot to contact them.}

% Tackling the manipulation tasks typically requires generating the reference signals and tracking these signals.

\begin{figure}[tb]
\begin{center}
	\includegraphics[width=3.2in]{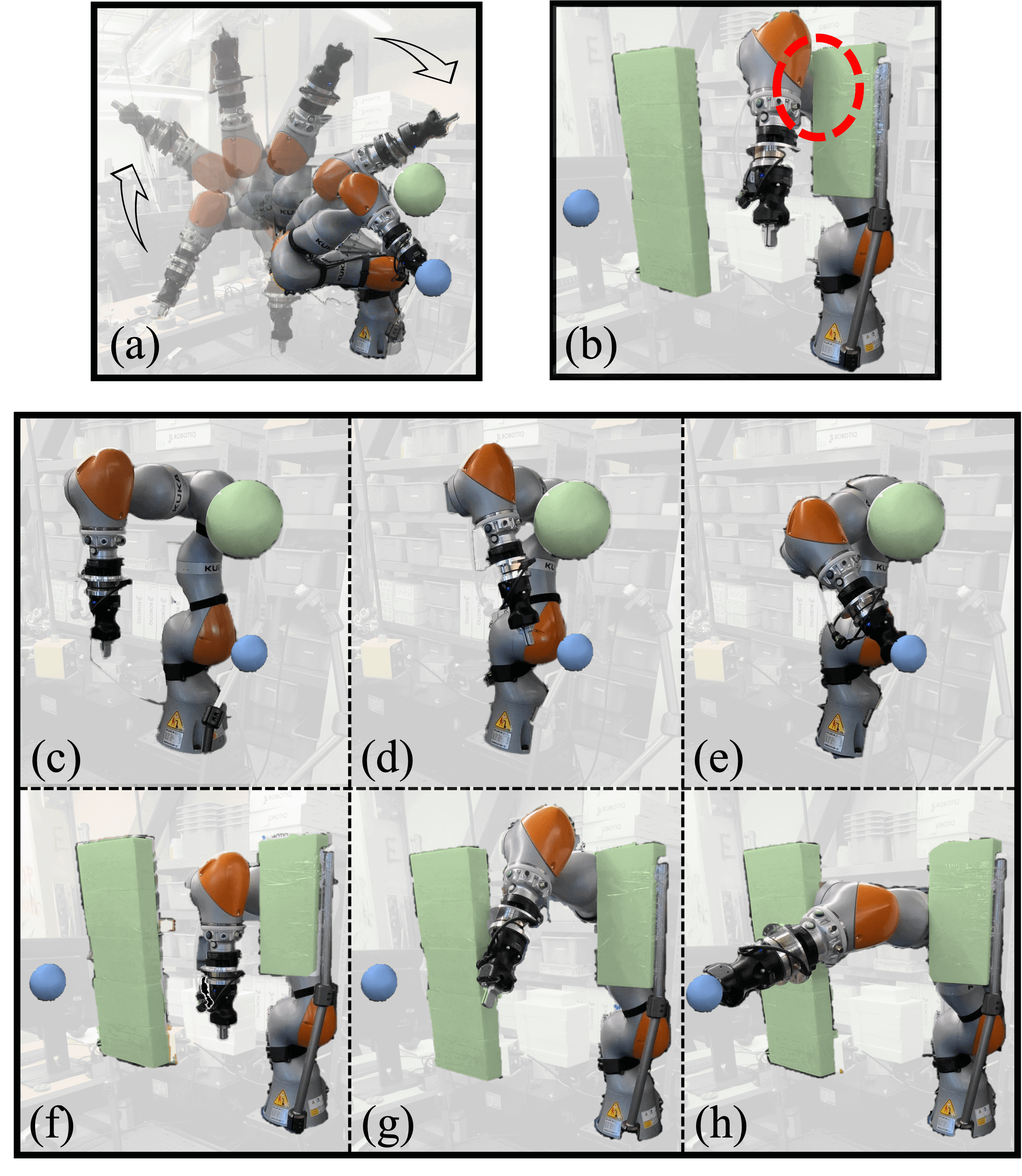}
% 	\vspace{-0.75em}
	\caption{\textbf{(a)} The robot finds a long and unnatural trajectory to reach the goal (blue) avoiding the obstacle (green). \textbf{(b)} The goal-reaching robot fails to find a collision-free trajectory; red circle indicates collision with the obstacle.
By allowing safe contact, the robot reaches the goal more efficiently \textbf{(c-e)}, and finds a trajectory in the highly collisional scenario \textbf{(f-h)}.}
	\label{fig: teaser}
\end{center}
\end{figure}

% \wenzhao{Something like In this work, we propose to plan motion allowing collisions, to emphasize this is a novel way proposed by us}
In this work, we propose to allow contact in robotic manipulation.
Allowing contact can enhance the robot's efficiency and capability by enlarging the feasible set~\cite{6906595}. The advantage of a larger feasible set is two-fold (Fig.~\ref{fig: teaser}). First, it yields better trajectories than enforcing collision-free constraints, measured by trajectory length and jerkiness. Second, it can generate feasible solutions in highly collisional scenarios where collision-free solutions cannot be found.

One of the most critical concerns in contact-allowed manipulation is safety. Without considering safety, the robot can hit obstacles heavily and damage the hardware. Safety constraints can be imposed by restricting the contact forces~\cite{1642072} and we impose this constraint by online generating and tracking of the reference signals.
For the prior, a constraint can be applied during trajectory optimization to limit the contact forces. For the latter, a robot controller can leverage null space compliance to reduce the contact force while tracking the desired trajectory~\cite{6094609}. 

% \wenzhao{Perhaps elaborate a bit, we are exploring the property that the DOFs are often higher than the task space DOF, like 7>6, or 6>3 not caring about rotation}
This paper studies the safe contact-allowed robotic goal-reaching problem with two feedback control loops. The outer loop optimizes the time-varying operational and joint trajectories in a receding horizon manner. The dynamics model of the robot and objects is approximated with a differentiable simulator, Brax~\cite{brax2021github}, and utilized to impose contact constraints.
The inner loop tracks the trajectory using an impedance controller.
This paper focuses on redundant manipulation where the robot is kinematically redundant to the task, and includes a null space projector in addition to an operational space control.
Compared with other works, our method optimizes joint configuration as extra decision variables and actively explores the null space for a safer motion. To solve the trajectory planning, we propose a hybrid algorithm that integrates sampling- and gradient-based methods, gaining the benefit of scalability and differentiability.

We empirically evaluate the proposed method in various simulation and real-world environments to demonstrate the effectiveness of the contact-allowed planning and control. In summary, our work makes the following contributions.
\begin{itemize}
  \item We state the contact-allowed robotic goal-reaching task with safety constraints. We provide open-source environments and benchmarks for the task.
%   \wenzhao{And provided a benchmark; we want to mention we plan to opensource the env and code}
%   These environments span different levels of collision conditions, from free space to highly collisional.
  \item We propose a time-varying trajectory planner and tracking controller for the contact-allowed problem. The planner optimizes both operational and null space reference signals to efficiently and safely achieve the goal.
  \item We present a hybrid optimization solver for the trajectory planner. The superiority of the proposed algorithm is evaluated in diverse experiments ranging from free space to highly collisional.
\end{itemize}

The remainder of the article is organized as follows. Related works are discussed in Chapter~\ref{sec: related_works}. The details of our method and environments are described in Chapter~\ref{sec: algorithms} and ~\ref{sec: env}. The experiments and conclusions are presented in Chapter~\ref{sec: exp} and~\ref{sec: conclusion}.

%%%%%%%%%%%%%%%%%%%%%%%%%%%%%%%%%%%%%%%%%%%%%%%%%%%%%%%%%%%%%%%%%%%%%%%%%%%%%
%%%%%%%%%%%%%%%%%%%%%%%%%%%%%%%% Related Works %%%%%%%%%%%%%%%%%%%%%%%%%%%%%%
%%%%%%%%%%%%%%%%%%%%%%%%%%%%%%%%%%%%%%%%%%%%%%%%%%%%%%%%%%%%%%%%%%%%%%%%%%%%%

\section{Related Work}
\label{sec: related_works}

\subsection{Impedance and Null Space Control}
% impedance control
Recently, problems and algorithms related to compliant behaviour have obtained an increasing amount of interest because of its capability of ensuring safe interaction.
Impedance control represents an appealing approach to achieve robot compliance with joint torque command~\cite{3140702, 1481160, vic}.
The impedance control uses a mass-spring-damper system to model the dynamic relationship between the robot and the contact objects. 
% The robot actuation torque is computed with a proportional-derivative (PD) controller to mimic the second-order dynamics.
% The operational impedance control for torque-controlled robots was investigated thoroughly in~\cite{, vic}. 
% Variable impedance control~\cite{vic} provides more flexibility than impedance control by also controlling the stiffness of the mass-spring-damper model.
Another method that focuses on increasing contact safety leverages kinematic redundancy with null space control~\cite{0278364908091463}.
% In~\cite{0278364908091463}, a theoretical and empirical evaluation of variant control techniques was presented for redundant manipulators.
% One approach to using redundant degrees is imposing null space control~\cite{0278364908091463}. 
In~\cite{5980228, 8793691, 6385690, 6684300}, the null space control was considered for a case where the joint space motion operates in the null space of the operational task.
% Null space impedance controls were proposed at joint acceleration and velocity levels with external torque observers~\cite{6094609, 6385690, 6684300}. 
These works showed it is effective to render desired joint impedance behaviour without affecting the operational task.
\cite{4651204, 8206331} models safety as the primary objective and optimizes the operational motion to diminish contact. In contrast, we optimize the joint and operational space motion simultaneously while preserving the operational task structure.
% \wenzhao{Please doublecheck the correctness of the rewrite}
% In ~\cite{4651204, 8206331}, ensuring safety is the primary objective, and the operational motion tries to diminish contact.
% In~\cite{8793691}, variant null space projectors were discussed to efficiently execute the operational task and reduce the contact by joint compensation. 
% This paper leverages a similar idea to design the robot controller: joint impedance motion is executed while preserving the operational task.

Receding horizon control strategies are typically utilized to generate and track reference signals with the above controller, such as model predictive control~\cite{0278364912456319, 24613282461380} or reinforcement learning~\cite{9811973, martin2019iros}. The former defines a cost function and control constraints to optimize a short-term motion. The latter selects actions based on the estimated reward or value at each time step. In~\cite{7750723}, the null space behaviour was optimized to avoid collisions while satisfying the task constraints. In~\cite{220713438}, a null space posture torque is applied to deal with the unexpected collision. This paper similarly leverages the receding horizon control framework to generate and track the trajectory. But unlike previous works, we treat the null space behaviour as an additional decision variable for the operational space motion, and optimize them at the same time for safe and efficient robotic manipulation.

\subsection{Contact-Allowed Robotic Manipulation}
Allowing safe contact between the robot and the environment can improve performance in specific tasks. Some works focus on contact-allowed trajectory planning with deformable obstacles. Different from collision-free problems, interactions between the robot and the environment impose uncertainties on the dynamics. In~\cite{6301045}, a simulator was included to deal with dynamical uncertainty. In~\cite{1642072}, the contact force and collision response were approximated with the finite-element method. Instead of approximating the dynamics, contact constraints were represented using voxel grids and intruding volumes in~\cite{6906595}. Although intruding distance avoids the usage of simulation, it only works for static obstacles where the object center stays fixed. This, however, is not true in practice since the obstacle can move along the contact direction. Moreover, parameterizing intruding volumes/distances is less explainable or intuitive than contact forces. Therefore, the above works only focus on finding feasible trajectories for simple deformable geometries. These trajectories are generated by the rapidly-exploring random tree (RRT) or its variants~\cite{rrt_star,28220132822036} in one shot and won't be updated during trajectory execution.
% \wenzhao{A bit confused on the last two sentences, they plan paths, but RRT is still required?}
% Ans: forgot to delete "then"

Other works have included robot control in contact-allowed manipulation using sample-based or optimization-based methods like control barrier functions~\cite{grandia2021multi}. These works, however, restrict the contact area to the feet for legged robot jogging~\cite{9561521, 9340745}, or the end effector for manipulator wiping and polishing~\cite{9145608, 9832727, ppojpo}. A comparison of control methods and task definitions was presented in~\cite{SUOMALAINEN2022104224}. Body contact is typically hard to sense and localize accurately~\cite{6094609}, limiting the usage of these controllers in the null space contact scenario. This work aims to allow contact between the full robot body and the environment for a goal-reaching task. We use simulation to approximate the system dynamics~\cite{1642072} and iteratively plan and execute robot motions~\cite{9145608}. 

% Depending on the uncertainty of the system dynamics, the robotic manipulation problem can be solved in a single shot or an iterative manner.
% For the prior, the whole motion is generated before moving, and the robot tracks the trajectory at execution. 
% The rapidly-exploring random tree (RRT) and its variants~\cite{rrt_star,28220132822036,Tahir_2018,9023003} are among the most popular in this category. These algorithms iteratively grow the connectivity tree until they arrive at the goal. 
% Gradient descent can also be combined in such a framework to optimize the trajectory locally.
% However, when the dynamics have uncertainty, one-shot planning can fail due to execution errors. 
% In this case, model predictive control~\cite{0278364912456319, 24613282461380} or reinforcement learning~\cite{9811973, martin2019iros} are preferred to generate a short-term motion based on the Markov decision process and repeat it at every time step.

% solver
% Evolutionary methods are one of the most popular methods to optimize the trajectory planning problem. These algorithms start from a broad set of candidates and evolve them towards more promising regions. 

%%%%%%%%%%%%%%%%%%%%%%%%%%%%%%%%%%%%%%%%%%%%%%%%%%%%%%%%%%%%%%%%%%%%%%%%%%%%
 %%%%%%%%%%%%%%%%%%%%%%%%%%%%%%%% Algorithms %%%%%%%%%%%%%%%%%%%%%%%%%%%%%%%
%%%%%%%%%%%%%%%%%%%%%%%%%%%%%%%%%%%%%%%%%%%%%%%%%%%%%%%%%%%%%%%%%%%%%%%%%%%%

\section{Contact-Allowed Robotic Goal-Reaching}
\label{sec: algorithms}

\subsection{Problem Formulation}
\label{subsec: formulation}
\begin{figure}[tb]
\begin{center}
	\includegraphics[width=3.2in]{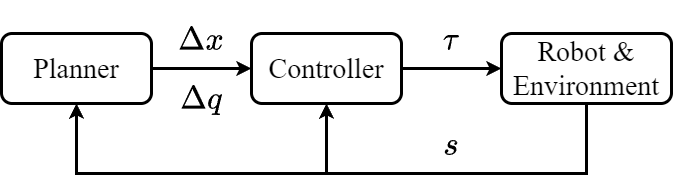}
% 	\vspace{-0.75em}
	\caption{The planning loop takes in state $s$ and optimizes reference signals (i.e., operational space references $\Delta x$ and null space references $\Delta q$) for the robot controller, which generates joint torques $\tau$ as actuation commands.}
	\label{fig: pipeline}
\end{center}
\end{figure}

This paper chooses goal-reaching as the operational manipulation task. It requires the end-effector to reach a given goal pose, while the manipulator can collide with the environment with a maximum permitted contact force. 
Relating to the formalisms in Chapter~\ref{sec: introduction},  a hierarchical framework is utilized to generate and track robotic motions, as shown in Fig.~\ref{fig: pipeline}. 
The outer planner takes the robotic and environmental state $s$ as input and optimizes operational and joint space motions $a=\{\Delta x, \Delta q\}$; the input includes robot joints and obstacle information\footnote{Poses for rigid obstacles; node positions for deformable obstacles represented by node graphs, as in Fig.~\ref{fig: envs}(c).}.
The inner robot controller computes joint torques $\tau$ based on the references to actuate the robot.
%  for efficient and safe manipulation
% \wenzhao{Introduce the notations here, like goal pose, reaching criteria something smaller than $\epsilon$, maximum force $\xi$, other metrics like torque, total energy can be left to future work}

The 3D goal pose is denoted as $s_g$. The success criteria is defined as $\left \| s_t - s_g \right \| \leq \delta$ ($\delta=0.01m$ in this paper), where $\left \| \cdot \right \|$ computes the translational distance from the end-effector to $s_g$. The system propagates with the transfer function $T(s_t, \tau_t)$ as the robot executes the torque command at $t$, which computes the successive state $s_{t+1}$ and the contact force. The maximum permitted contact force is written as $\varepsilon$.

\subsection{Operational and Null Space Impedance Control}
\label{subsec: controller}
% Introduce the low-level control
In this work, we adopt operational space computed torque control~\cite{3140702} for the manipulation task and introduce null space projection~\cite{6094609} for safe contact.
Previous works have shown the advantage of planning task-related motions in the operational space over the joint space~\cite{martin2019iros}. Not only the manipulation tasks are typically defined in the operational space, but also it is straightforward to adapt the controller among different robots.
Moreover, null space projection allows compliance at the joint level for a redundant robot without disturbing the operational task, thus improving safety by reducing contact forces.

This work uses the computed torque control~\cite{206book} as the operational space controller. It first calculates the wrench for the end-effector based on the desired motion and then maps the wrench to joint torques. The control law is written as (\ref{eq: control_operational}).
\begin{equation}
\label{eq: control_operational}
    \tau_{op} = J(q)^T \Lambda(q) \left ( -K_p \Delta x - K_d \dot{x} \right ) + C(q,\dot{q}) + g(q)
\end{equation}

For a robotic manipulator with $n$ joints, $q \in \mathbb{R}^n$ is the joint angle, $x\in \mathbb{SE}(3)$ is the end-effector pose, $J(q)$ is the task Jacobian, $\Lambda(q) = (J(q) M^{-1}(q) J(q)^T)^{-1}$ is the operational space inertial matrix and $M(q)\in \mathbb{R}^{n\times n}$ is the robot inertial matrix, $C(q,\dot{q})\in \mathbb{R}^n$ is the Coriolis and Centrifugal forces, $g(q)\in \mathbb{R}^n$ is the robot gravity vector.
We use $\Delta x = x \ominus x_d$ to represent the error in $\mathbb{SE}(3)$, where $x_d$ is a desired operational pose. $K_p, K_d \succeq 0$ are PD gains for the controller. The following sections will drop the dependencies in $q, \dot{q}$ for better readability.

The null space control projects joint torques into the null space with the projector $N = I - J^{\dagger} J$, where $J^{\dagger} = M^{-1} J^T \Lambda$ is the dynamically consistent inverse of the Jacobian~\cite{8793691, 220713438}. The projected torque does not affect the motion in the operational space.
% One can see that for a joint torque $\tau$ and the projected torque $\tau_p=N^T \tau$, $(J^{T})^{\dagger} \tau_p = 0$, resulting in zero wrench for the end-effector.
We implement a joint space PD controller to generate and project the torque:
\begin{equation} 
\label{eq: control_null}
    \tau_{null} = N^T \left( -K_{qp} \Delta q - K_{qd} \dot{q} \right )
\end{equation}
, where $\Delta q = q-q_d$ and $q_d$ is a desired joint posture. $K_{qp}, K_{qd} \succeq 0$ are control gains in the joint space.

Combining the operational space control (\ref{eq: control_operational}) with the null space projection (\ref{eq: control_null}), the final control output is defined as:
\begin{equation} 
\label{eq: control_all}
    \tau = \tau_{op} + \tau_{null}.
\end{equation}

\subsection{Contact-Allowed Motion Planning}
% \wenzhao{Change to something like "collision allowed motion planning"? Section A introduced the controller, not much more info with the current title.}
\label{subsec: traj_plan}
% Introduce the high-level planner: opt. formulation, simulation, solver
Section~\ref{subsec: controller} introduces an operational space controller with null space compliance. The control law (\ref{eq: control_operational},~\ref{eq: control_null},~\ref{eq: control_all}) takes in the reference signals $a=\{\Delta x, \Delta q\}$ to compute the actuation torque. This section introduces how to find the time-varying reference signals for efficient and safe robotic manipulation.
% and control gains $\{K_p, K_d, K_{qp}, K_{qd}\}$ 

The problem is formulated in a receding horizon manner. For each time step $k$, the optimization is written as follows.
\begin{subequations}
\label{eq: opt}
\begin{align}
\min_{a_{k:k+H-1}} & \sum_{t=k}^{k+H-1} \lambda_1 \left \| s_{t+1} - s_g \right \| - \lambda_2 \left \| s_{t+1} - s_k \right \| \label{eq: opt_obj}\\
\textrm{s.t.} \quad  & \enspace \tau_t = f(a_t) \qquad \triangleright~(\ref{eq: control_all}) \label{eq: opt_c1} \\
& \enspace s_{t+1} = T(s_t, \tau_t) \label{eq: opt_c2} \\
& \enspace \textrm{contact}(s_t, a_t) \leq \varepsilon \label{eq: opt_c3} \\
& \enspace \qquad \textrm{for} \ t\in [k,..., k+H-1] \nonumber
\end{align}
\end{subequations}
% We use the subscript $t$ to represent different time steps. Control inputs $a=\{\Delta x, \Delta q\}$ are abbreviated and regarded as decision variables.

The objective function~(\ref{eq: opt_obj}) minimizes costs over the horizon $H$. 
% $s_{t+1}$ is the successive state after applying $a_t$ at $s_t$. $s_g$ is the predefined goal state.
$s_k$ is the initial state at the current planning horizon. $\lambda_1, \lambda_2>0$ are hyper-parameters.
$\left \| \cdot \right \|$ measures the distance of end-effectors between states.
The first term minimizes the distance to the goal, representing the objective to reach the destination efficiently. The second term encourages exploration by forcing the robot away from the initial state. In practice, it helps to escape from the local minimum.

Constraint~(\ref{eq: opt_c1}) stands for the control law in~(\ref{eq: control_all}). Constraint~(\ref{eq: opt_c2}) represents the transfer function.
% The approximation of the transfer function will be discussed in the next section. 
Constraint~(\ref{eq: opt_c3}) encodes the safety requirement, where $\textrm{contact}(s_t, a_t)$ measures the contact force while executing the robot control; the contact force is constrained by the maximum bound $\varepsilon$.

The transfer function $T(s_t, \tau_t)$ approximates the system propagation in~(\ref{eq: opt_c2}) and estimates the contact force in~(\ref{eq: opt_c3}).
% \wenzhao{Is it fair to say, one insight/discovery is the $T$ does not need to be accurate. So we choose Brax with quite rough approximate physics models}
% The dynamics contain the robot motion~(\ref{eq: operation_dynamics}), the robot-environment interaction due to contacts, and the environmental motion.
We use the Brax physics engine~\cite{brax2021github} as the transfer function. Brax is designed for performance and parallelism on accelerators, allowing a large sample size for sampling-based optimization solvers. Moreover, it supports auto-differentiation of the dynamics, which makes it possible to use gradient-based solvers. The next section introduces how we solve the optimization problem by leveraging the parallelism and differentiable properties of the Brax engine.

\subsection{Solving the Optimization Problem}
\label{subsec: solver}
The optimization~(\ref{eq: opt}) is first written in a non-constraint form by plugging dynamics~(\ref{eq: opt_c1}, \ref{eq: opt_c2}) into the objective~(\ref{eq: opt_obj}) and simplifying contact constraints~(\ref{eq: opt_c3}):
\begin{equation} 
\begin{aligned}
\label{eq: opt_no_st}
    \min_{a_{k:k+H-1}} \sum_{t=k}^{k+H-1} & \lambda_1 \left \| T(s_t, f(a_t)) - s_g \right \| \\ 
- \enspace & \lambda_2 \left \| T(s_t, f(a_t)) - s_k \right \| \\ 
+ \enspace & \lambda_3 \left ( \textrm{contact}(s_t, a_t) - \varepsilon \right )
\end{aligned}
\end{equation}

To solve~(\ref{eq: opt_no_st}), this paper proposes a hybrid solver that integrates the covariance matrix adaptation evolution strategy (CMA-ES,~\cite{cmaes}) and the gradient descent method, as outlined in Algorithm~\ref{algo: solver}. The cost function in~(\ref{eq: opt_no_st}) is denoted as $\mathcal{L}$.
% For each time step, only the first action is selected for execution as a receding horizon control.

\begin{algorithm}
   \caption{Hybrid Optimization Solver}
   \label{algo: solver}
    \begin{algorithmic}[1]
    %   \Function{$\min \mathcal{L}$}{$a_{k:k+H-1}$} \Comment{(\ref{eq: opt_no_st})}
        \State \textbf{Initialize} $m=0, \ \sigma=0.01, \ C=I, \ p_{\sigma}=p_{c}=0$
        \State $k=200, \ k_{top}=50$ \Comment{Population size} \label{eq: solver_pop_size}
        \State $\gamma, \ \beta_{max}=1$ \Comment{Boundaries \& max grad. step}
        \For{$step=1$ to $max \ step$} 
            \For{$i \in \{ 1,..., k \}$} \label{eq: solver_cmaes_ask}
                \State $a_i \sim \mathcal{N}(m, \sigma^2 C)$ \Comment{Sample candidates}
                \State $a_{r} = \textrm{clip}(a_i, -\gamma, \gamma)$ \label{eq: clip} \Comment{Repair candidates}
                \State $d_i = \left \| a_i - a_r \right \|^2$
                \State $\alpha = \left\{\begin{matrix}
\exp({\log \mathcal{L}(a_i) - \log d_i}) & \textrm{if} \enspace d_i > 0 \\ 
0 &  \textrm{otherwise}
\end{matrix}\right.$ \label{eq: solver_alpha}
                \State $l_i = \mathcal{L}(a_r) + \alpha\cdot d_i$ \Comment{(\ref{eq: opt_no_st}) w/ penalty} \label{eq: solver_penalty}
            \EndFor
            \State $a_{best} = \textrm{argsort}(l_{1:k})$ \Comment{Find $k_{top}$ best candidates}
            \State $m = \textrm{mean}(a_{best})$ \Comment{CMA-ES mean}
            % \For{$a \in \{a_{1},..., a_{k_{top}}\}$} \Comment{Grad. descent}
                \State $\nabla \mathcal{L}=\frac{\partial \mathcal{L}(m)}{\partial a}$ \Comment{Diff. w/ Brax} \label{eq: grad_start}
                \State $\beta_1,..., \beta_k \sim \mathcal{U}[0, \beta_{max}]$
                % \State $\beta=\argmin_{\beta_{1:k}} \mathcal{L}(m + \beta_j \cdot \nabla \mathcal{L})$ \Comment{Grad. step size}
                \State $m = m + \argmin_{\beta_{1:k}} \mathcal{L}(m + \beta_j \cdot \nabla \mathcal{L}) \cdot \nabla \mathcal{L}$ \label{eq: grad_end}
                \State $m = \textrm{clip}(m, -\gamma, \gamma)$
            % \EndFor
            \State Update $\sigma, \ C, \ p_{\sigma}, \ p_{c}$ \Comment{CMA-ES update~\cite{cmaes}}
        \EndFor
\end{algorithmic}
\end{algorithm}

Compared to others~\cite{cmaes, 220700167}, our method uses a larger population size, bounds the decision variables, and applies an additional gradient descent step. The boundary constrains the robot's movement in operational and joint space. A penalty is added to the cost function, which is the distance to the boundary. A tradeoff weight $\alpha$ is adaptively selected so that the cost and the penalty are similar in magnitude.
% The logarithms of the cost and the distance are used to estimate the magnitude difference.
% \wenzhao{Is it possible to elaborate a bit, here or in Exp?}.
Moreover, we add a gradient descent step before refitting distributions in CMA-ES. The gradient step optimizes the solution locally. Similar ideas have been utilized in~\cite{220700167, rika_33} and have shown improved performance over CMA-ES alone. Nevertheless, our method does not iterate the gradient descent until convergence.
% \wenzhao{elaborate the rational of single step iteration}. 
We sample a population of neighbours in the gradient direction and find the best one as the local optimal, similar to~\cite{220700167}. 
Such design gains an advantage in computation speed as the evaluation of the cost is much faster than that of the gradient. Thus, we sample more candidates and only apply the gradient descent once per step.
The algorithm is halted if the maximum step has been reached or the cost stagnates for three steps. 
Readers are referred to~\cite{cmaes} for details of the CMA-ES refitting.

% In practice, we evaluate the cost with the Brax in parallel. $k$ samples are evaluated in the same batch. The computation of the cost is 10 times faster than that of the gradient. Thus, more forward evaluations are used in Algorithm~\ref{algo: solver} than backward gradients.  

%%%%%%%%%%%%%%%%%%%%%%%%%%%%%%%%%%%%%%%%%%%%%%%%%%%%%%%%%%%%%%%%%%%%%%%%%%%%%
%%%%%%%%%%%%%%%%%%%%%%%%%%%%%% Environments %%%%%%%%%%%%%%%%%%%%%%%%%%%%%%%%%
%%%%%%%%%%%%%%%%%%%%%%%%%%%%%%%%%%%%%%%%%%%%%%%%%%%%%%%%%%%%%%%%%%%%%%%%%%%%%

\begin{figure}[t]
\begin{center}
	\includegraphics[width=3.1in]{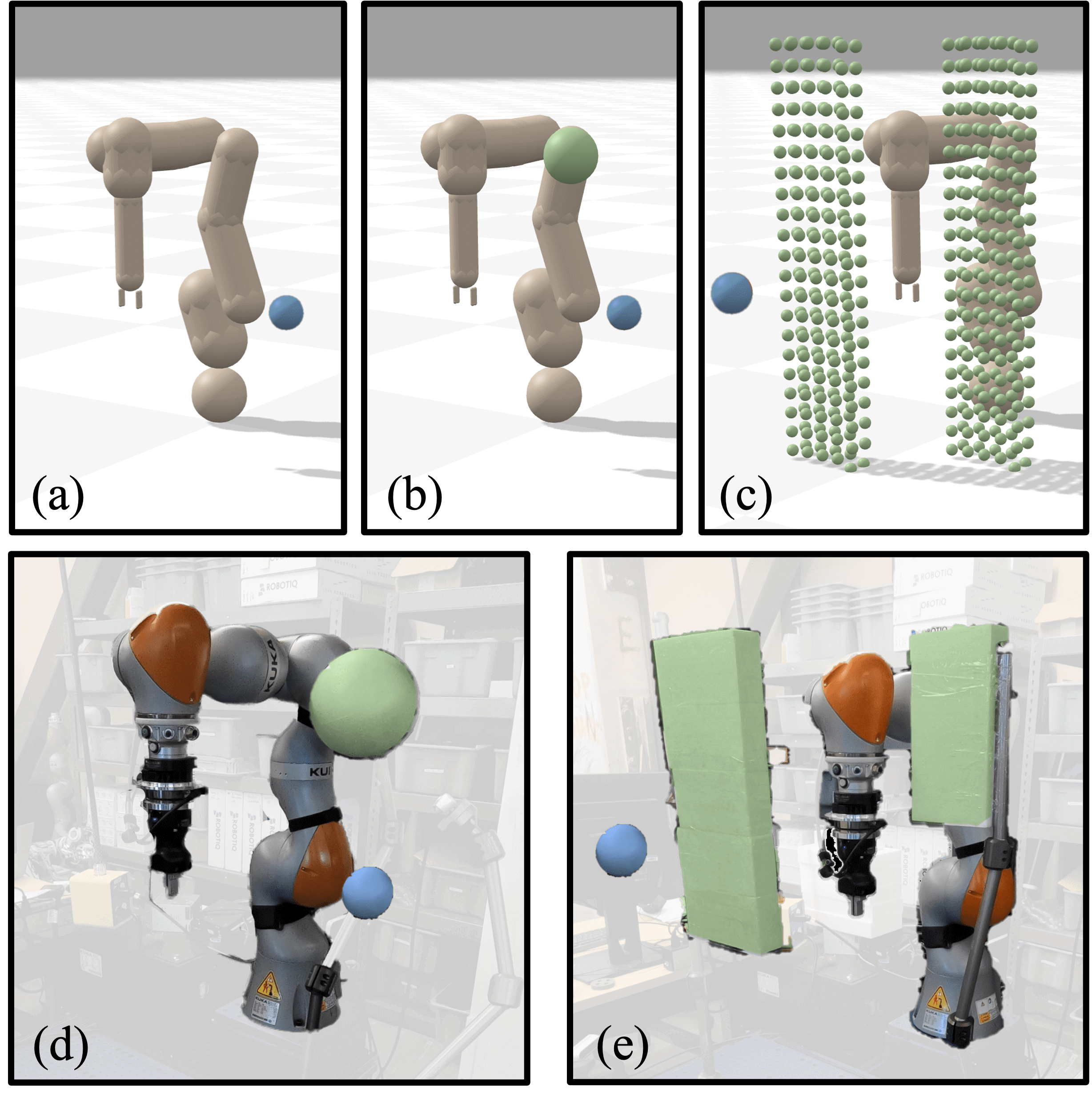}
% 	\vspace{-0.75em}
	\caption{Goal-reaching task environments of different collision complexity levels. The target is blue, and the obstacles are green. (a) Free space, (b) ball obstacle, (c) wall obstacle, (d) real-world ball obstacle, and (e) real-world wall obstacle.}
	\label{fig: envs}
\end{center}
\end{figure}

\section{Environments}
\label{sec: env}

% The previous chapter illustrates the formulation of the goal-reaching task, compliance null space controller, motion planning optimization, and the solver. 
This chapter introduces the simulation and the real-world environments in which we evaluate and compare different control and planning methods, as visualized in Fig.~\ref{fig: envs}.
These environments span different levels of collision conditions, ranging from free space to highly collisional.
% \wenzhao{outdated: The obstacles include both rigid and deformable objects. We also add a strawberry picking task to demonstrate the practical usage of the contact-allowed problem.}

The simuation environments are built with Brax~\cite{brax2021github}, where the robot is modelled with cylinder links to imitate the kinematics and dynamics of a real one (i.e., Kuka iiwa 14). Such modelling significantly reduces the computation time with simpler contact dynamics. The following sections introduce the detailed configurations for each environment.

\subsection{Free Space}
\label{subsec: free_space}
This environment aims to measure the performance of the proposed method in a collision-free condition. 
The target is randomly placed inside the robot manipulability ellipsoid. 
Since there is no obstacle, the optimal trajectory in the operational space is a straight line, as defined in~(\ref{eq: opt}, \ref{eq: opt_no_st}).
% We measure the time for the robot to arrive at the goal.

\subsection{Ball Obstacle}
\label{subsec: ball_obstacle}
% \wenzhao{as discussed, rigid and deformable can be ambiguous. Either way, ball and wall in subsection C should be the same.}
This environment intends to show that it is more efficient to accomplish certain manipulation tasks by allowing safe contacts.
In this environment, a ball with $0.1m$ radius is added as an obstacle. The target is randomly placed within a workspace,
%where most robotic manipulation tasks are defined: 
varying from $[0.2, -0.3, 0.0]$ to $[0.6, 0.3, 0.5]$ in $[x, y, z]$, respectively. The obstacle is placed to collide with the robot in the free space trajectory. Specifically, the optimal path is first computed without the obstacle. Then the ball is added to hamper the optimal path by colliding with the robot's middle bodies. 
Although the obstacle makes the free space trajectory infeasible, other collision-free paths exist. 

\subsection{Wall Obstacle}
\label{subsec: wall_obstacle}
This environment demonstrates that allowing safe contacts enables highly constrained manipulation tasks where collision-free paths cannot be found~\cite{1642072}.
The environment puts two $1\times 0.1\times 0.2m$ walls between the robot and the target. The walls obstruct all collision-free paths. 

The deformable wall is modelled as a mass-spring system in the Brax simulator. The wall is tessellated to volumetric finite element meshes, including vertices and edges. We use small rigid spheres to represent the mesh vertices and spring joints to represent the mesh edges. The simulation can mimic the physical deformation by carefully parameterizing the sphere mass and the spring stiffness. The robot's initial joint configuration is fixed through trials, while the target pose is randomized in front of the robot.

% \subsection{Strawberry Picking}
% \label{subsec: strawberry_pick}
% % \wenzhao{It might be helpful to have a one sentence description emphasizing the environment purpose for each subsection. For example for strawberry, we aim to evaluate how the planners handle complex deformable environments in a realistic setting.}
% This environment evaluates how algorithms handle complex deformable environments in a realistic setting.
% The goal is to pick strawberries from the plant. The strawberry plant has multiple petioles, which grow trifoliate leaves or strawberry fruits. Since detecting and localising the strawberry is not the main focus of this paper, we assume the fruits are well detected. Moreover, the trifoliate leaves and petioles are considered obstacles when picking the fruit. 

% The structure of the strawberry plant is randomly generated based on the botanical topology. A fixed spiral phyllotaxis arranges strawberry petioles, whereby every fifth leaf is situated exactly above the first one. Each plant has 3 to 20 petioles, and each petiole is 0.1 to 0.4$m$ long.

% The deformable strawberry plant is modelled with rigid physics in the simulation. The flexible petiole and leaf are connected cylinders with spherical joints and appropriate stiffness and damping. The fruits are spheres attached to the end of the petiole, as the image shows.
% We also generate a real strawberry picking environment with a plastic plant. The fruits and leaves are detected with RGB and depth cameras.

\subsection{Real-World}
\label{subsec: real_world}

Besides the simulation, we create physical environments with foam balls and walls to evaluate the proposed method in the real world. The foam ball has a $0.1m$ radius, whose position is generated similarly as in simulation. The foam walls are positioned with fixtures as shown in Fig.~\ref{fig: envs}(e), and have the same effective dimensions as in simulation. Since tracking the real-world system is not the main focus of this work, we assume the position and deformation of the foam obstacles are known. We manually tuned the coefficients in the dynamics approximator (i.e., Brax) to obtain an accurate transfer function~(\ref{eq: opt_c2}). In practice, the transfer function and the system state can be estimated with an additional observer, as suggested in~\cite{9812092, zhu2022synthesizing}. 

%%%%%%%%%%%%%%%%%%%%%%%%%%%%%%%%%%%%%%%%%%%%%%%%%%%%%%%%%%%%%%%%%%%%%%%%%%%%%
%%%%%%%%%%%%%%%%%%%%%%%%%%%%%% Experiments %%%%%%%%%%%%%%%%%%%%%%%%%%%%%%%%%%
%%%%%%%%%%%%%%%%%%%%%%%%%%%%%%%%%%%%%%%%%%%%%%%%%%%%%%%%%%%%%%%%%%%%%%%%%%%%%

\section{Experiments}
\label{sec: exp}

\begin{table*}[]
\centering
\caption{Comparison of trajectory efficiency and safety metrics in simulated and real-world environments. In each cell, the first element is the task execution time ($s$); the second is the maximum contact force on the robot body in simulation ($N$) or the maximum computed external torque on robot joints in real-world ($Nm$).}
% \wenzhao{1. It looks suspicious real and sim are exactly the same in run time. Forgot to update? 2. For free-space, is it exactly the same linear motion? 3. The torque values in real-world seem quite large, is it correct?}
\label{tab1}
\begin{tabular}{c||cc|cc|cc||cc|cc}
                        & \multicolumn{2}{c|}{\textbf{Free Space}} & \multicolumn{2}{c|}{\textbf{Ball Obstacle}} & \multicolumn{2}{c||}{\textbf{Wall Obstacle}} & \multicolumn{2}{c|}{\textbf{Ball Obstacle Real}} & \multicolumn{2}{c}{\textbf{Wall Obstacle Real}} \\ \hline\hline
\textbf{Collision-Free} & $11.4\pm 1.4$               & $0$               & $37.8\pm 3.9$              & $0$                   & \multicolumn{2}{c||}{Fail}                   & $38.4\pm 7.9$                  & $0$                    & \multicolumn{2}{c}{Fail}                        \\
\textbf{Ref. Posture}   & $11.4\pm 1.4$               & $0$               & $13.1\pm 0.4$              & $4.7\pm 2.3$             & $21.3\pm 2.0$              & $1.9\pm 0.7$             & $15.2\pm 1.7$                  & $1.6\pm 0.2$                  & $18.7\pm 1.1$                  & $1.1\pm 0.1$                 \\
\textbf{Ours}           & $11.4\pm 1.6$               & $0$               & $12.8\pm 0.5$              & $2.8\pm 1.2$             & $17.5\pm 1.5$              & $0.5\pm 0.3$             & $14.7\pm 1.6$                  & $0.7\pm 0.1$                  & $15.0\pm 1.0$                  & $0.5\pm 0.1$                
\end{tabular}
\end{table*}

This chapter provides experiments for the proposed method in environments described in Chapter~\ref{sec: env}.
The goal of the experiments is three-fold. First, we demonstrate the advantage of allowing contacts in multiple collisional scenarios. Second, we show the benefit of generating and tracking both the operational and null space reference signals. Third, we provide empirical evaluations for the proposed hybrid optimization solver.

The hyperparameters used throughout our experiments are set as follows.
$\lambda_{1,2,3}=\{1, 0.2, 5\}$, $H=3$, $\varepsilon=10N$, $max \ step=100$, $\gamma=[0.01m, 0.2rad, 0.2rad]$ where the first two represent the end-effector maximum movement and the last limits the null space motion.
In this paper, we do not infer control gains; instead, they are set to constants: $K_p=880I, K_d=100I, K_{qp}=30I, K_{qd}=I$. On the one hand, gains work as a scaling factor for $\Delta x, \Delta q$ in~(\ref{eq: control_all}), thus implicitly optimized. We did not observe improvement by including gain to the decision variables in our experiments. On the other hand, including gain inference makes the algorithm less stable and harder to solve. The control torque is prone to going unbounded with the inferred scaling.

\subsection{Collision-Free or Contact-Allowed}
\label{subsec: baseline}

\begin{figure}[tb]
\begin{center}
	\includegraphics[width=3.4in]{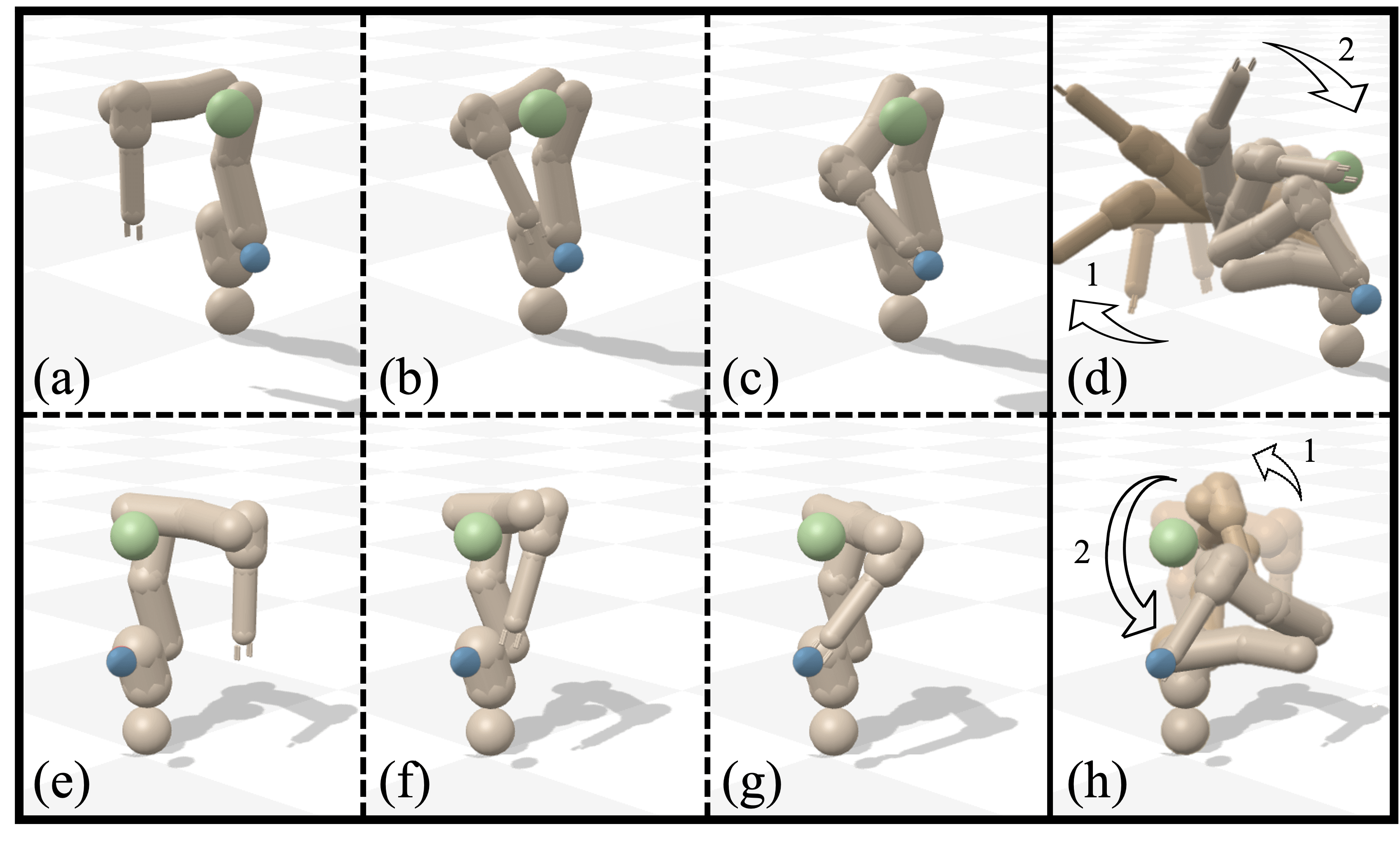}
% 	\vspace{-0.75em}
	\caption{Execution trajectories in simulated ball environments. (a-c, e-g) show two trajectories generated by our method, (d, h) show the collision-free trajectories with the same environmental settings.}
	\label{fig: rrt_vs_ours}
\end{center}
\end{figure}

This experiment compares the performance of our contact-allowed planner with a collision-free trajectory planner. We used the RRT$^*$~\cite{rrt_star} to generate the collision-free trajectory in the joint space and our controller to track the path without replanning. Each experiment was repeated three times to record the average task completion time in Table~\ref{tab1}. The maximum contact force throughout the execution is reported in simulated environments. A torque observer was implemented to measure the external torque in the real world~\cite{220713438}. 

For contact-allowed goal-reaching, an alternative to the receding horizon planning is to plan the whole trajectory before moving the robot~\cite{6301045, 1642072}.
However, the large search space, from allowing contact and additional null space motion, makes the planning problem intractable. Thus, we do not include such a planning framework in the comparison.
% However, the empirical implementations do not offer tractable computation time. The contact allowance and additional null space motions enlarge the search space. Thus, we do not include such a planning framework in the comparison.

As seen in the ball obstacle experiment in Table~\ref{tab1}, allowing safe contact improves the task efficiency with the ball obstacle. Although contact forces appeared during the execution, our method reduces the task completion time by 2.6\textbf{x}. Fig.~\ref{fig: rrt_vs_ours} visualizes two example executions in the simulated environment. In the highly collisional environment, i.e., wall obstacles, a collision-free trajectory cannot be found to reach the target. In contrast, our method completes the task by pushing the obstacles away. 
These results indicate that allowing collision relaxes the optimization constraints and provides a larger feasible motion set.
% Thus, a better trajectory can be discovered in narrow problems.
Lastly, in the free-space environment comparison, we observed the same performance, suggesting that allowing contact does not change the behaviour in unconstrained scenarios.

\subsection{Different Control Laws and Decision Variables}
\label{subsec: ablation}

\begin{figure}[tb]
\begin{center}
	\includegraphics[width=3.3in]{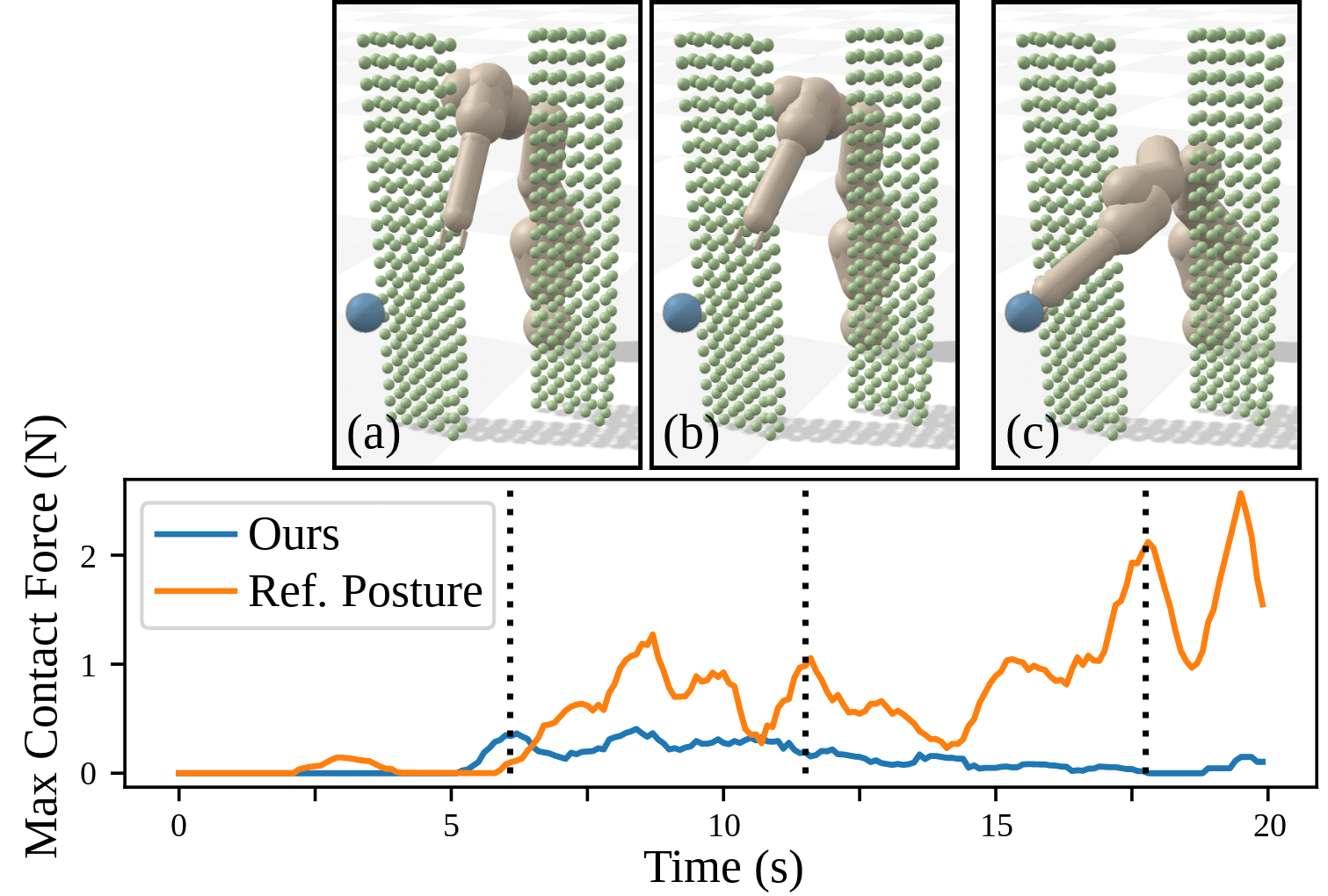}
% 	\vspace{-0.75em}
	\caption{Example contact force profile when reaching a goal in the wall environment. Pictures correspond to the robot configuration induced by our method at time stamps marked by dotted vertical lines; (a, b) the robot tracks trajectories to push obstacles in a compliant manner and adjusts its joint configuration in the null space; (c) the robot reaches the goal while maintaining a minimum contact force. Compared to the ablation method, controlling both operational and null space trajectories reduces the overall contact forces.}
	\label{fig: wall_traj}
\end{center}
\end{figure}

This experiment provides ablation studies to the control laws and decision variables in~(\ref{eq: opt}).
The ablation method applies a reference posture torque in the null space to track a joint reference. We set the desired joint posture $q_d=0$ in (\ref{eq: control_null}).
% representing that the robot's configuration should be close to the initial pose. 
The control law is written as:
\begin{equation} 
\label{eq: abalaton_2}
    \tau_{posture} = \tau_{op} + N^T \left ( -K_{qp} q - K_{qd}\dot{q} \right )
\end{equation}
The trajectory planner only optimizes the operational space motion $\Delta x$ and has no control over the null space.

Based on the results in Table~\ref{tab1}, our method outperforms the reference posture control method in all environments. Fig.~\ref{fig: wall_traj} shows the contact force profile for an experiment in the wall environment. Our method reduces the contact force by almost 4\textbf{x} in the simulated wall environments and 1.7\textbf{x} in the ball environments. Our method also reduces the task execution time by generating shorter operational trajectories. The real-world experiments further support the findings that considering additional null space motions increases task safety. These results validate the necessity of optimizing the null space motion.
% We hypothesize that such observation is due to a larger search space. 
Since the null space behaviour is neither optimized nor controlled in the ablation method, it has a tighter search space than ours and often cannot find a trajectory with the same performance as ours.

Another way to plan robot trajectories is to optimize the motor torque directly. We implemented a joint torque planner and observed a similar performance to ours.
In contrast, our method maps the planned trajectory to motor torque with a control law defined in (\ref{eq: control_all}). Such structured design can be considered a special case of directly torque planning.
% Our method transfers the trajectory to torque with a control law. Such mapping might be considered within the optimization of torques, thus generating similar robot motions.
Nevertheless, optimizing joint torques directly induces a more complex optimization problem and requires more computation to solve. For example, in the wall obstacle environment, our method takes on average $0.47s$ to converge for each step, while planning torque directly takes $1.34s$. Similar results have been recognized in~\cite{martin2019iros}.
Moreover, since manipulation tasks such as our considered goal-reaching are often defined in the operational space, it is more intuitive and explicit to search for actions in the operational space. 
% Since the goal-reaching task is defined in the operational space, it might be more explicit to search for operational action than torque command. Moreover, it shows that the proposed control law reduces the optimization complexity while maintaining a similar solution set.

\subsection{With or Without Gradient Descent}
\label{subsec: compare_solver}
This experiment analyzes the proposed hybrid optimization solver in the wall obstacle environment.
We demonstrate the effectiveness of the bounded CMA-ES and single-step gradient descent, line \ref{eq: clip}-\ref{eq: solver_alpha} and \ref{eq: grad_end} in Algorithm \ref{algo: solver}.
% We demonstrate the effectiveness of using single-step gradient descent, line~\ref{eq: grad_end} in Algorithm~\ref{algo: solver}, with the bounded CMA-ES.
Two baselines were used for comparison. The first only uses the vanilla CMA-ES to search for the optimal reference signals~\cite{cmaes}, while the second involves multiple gradient steps until convergence, similar to~\cite{220700167}.

Interestingly, all solvers yield similar trajectories given the same environmental configuration, suggesting that they have converged to close solutions. Thus, we use the average cost decrease after 10 optimization steps as the comparison metric.
Results showed that our method ($0.34$) outperforms the vanilla CMA-ES ($0.22$). This validates the benefit of integrating gradient descent with the sampling-based approach; it expedites the convergence by locally refining the solution. Meanwhile, we observed further improved performance with multiple gradient steps ($0.40$). However, the iterative gradient calculation requires much longer computation time and significantly reduces the planning frequency. With the consideration of real-time performance, we applied single-step gradient descent in our proposed solver.

%%%%%%%%%%%%%%%%%%%%%%%%%%%%%%%%%%%%%%%%%%%%%%%%%%%%%%%%%%%%%%%%%%%%%%%%%%%%%
%%%%%%%%%%%%%%%%%%%%%%%%%%%%%%%% Conclusion %%%%%%%%%%%%%%%%%%%%%%%%%%%%%%%%%
%%%%%%%%%%%%%%%%%%%%%%%%%%%%%%%%%%%%%%%%%%%%%%%%%%%%%%%%%%%%%%%%%%%%%%%%%%%%%

\section{Conclusion}
\label{sec: conclusion}

This paper investigates contact-allowed robotic goal-reaching with operational and null space control. Our work has several key contributions.
First, we state the contact-allowed robotic manipulation problem with safety constraints, and provide a set of open-source environments for contact-allowed goal-reaching. These environments have different collision conditions, from free space to highly collisional, where collision-free solutions do not exist. Second, to generate and track reference signals for collision-allowed motion, we propose a receding horizon trajectory planner that optimizes the operational and null space reference signals, and track the reference with an impedance controller. Lastly, we present a hybrid solver to optimize reference signals. Simulation and real-world experiments indicate that using the proposed algorithm and solver, by allowing contact, our method achieves the manipulation goal efficiently and safely.
% First, we state the contact-allowed robotic manipulation task with safety constraints. Second, we provide open-source simulated and real-world environments for the contact-allowed problem. These environments have different collision conditions, from free space to highly collisional, where collision-free solutions do not exist. Third, we propose an algorithm to generate and track reference signals for the task; The operational and null space references are optimized in a receding horizon trajectory planner and tracked with an impedance controller. Lastly, we present a hybrid solver to optimize reference signals. Using the proposed algorithm and solver, our method can efficiently achieve the manipulation goal by allowing contact and can improve safety by planning both operational and null space behavior, as indicated by the experimental results.

The present work has some limitations. First, we do not demonstrate a more complex operational task than goal-reaching. It is interesting to generalize the optimization objective to other tasks by combining reinforcement learning or model predictive control techniques. Second, the current algorithm assumes known environmental states. There is no state observer during the real-world execution and thus is susceptible to sim-to-real gaps. In future works, we plan to address these limitations and develop contact-allowed algorithms for diverse robotic manipulation tasks, such as assembly~\cite{lian2021benchmarking}, table wiping, bin-picking~\cite{9812092, 9811685}, and other human-robot interaction tasks.
% The optimization objective, however, can be generalized to other tasks by combining reinforcement learning or learning model predictive control techniques~\cite{ppojpo}.
% Second, the current algorithm is an open-loop system for real-world experiments. There is no feedback to modify the trajectory online. In our future work, we will focus on addressing these limitations and developing contact-allowed algorithms for extensive robotic manipulation tasks, such as peg-in-hole insertion, table wiping, bin-picking~\cite{9812092, 9811685} with unexpected human interactions, and other human-robot interaction tasks.

%%%%%%%%%%%%%%%%%%%%%%%%%%%%%%%%%%%%%%%%%%%%%%%%%%%%%%%%%%%%%%%%%%%%%%%%%%%%%%%%

\addtolength{\textheight}{-1cm}   % This command serves to balance the column lengths
                                  % on the last page of the document manually. It shortens
                                  % the textheight of the last page by a suitable amount.
                                  % This command does not take effect until the next page
                                  % so it should come on the page before the last. Make
                                  % sure that you do not shorten the textheight too much.

%%%%%%%%%%%%%%%%%%%%%%%%%%%%%%%%%%%%%%%%%%%%%%%%%%%%%%%%%%%%%%%%%%%%%%%%%%%%%%%%

%%%%%%%%%%%%%%%%%%%%%%%%%%%%%%%%%%%%%%%%%%%%%%%%%%%%%%%%%%%%%%%%%%%%%%%%%%%%%%%%

%%%%%%%%%%%%%%%%%%%%%%%%%%%%%%%%%%%%%%%%%%%%%%%%%%%%%%%%%%%%%%%%%%%%%%%%%%%%%%%%
\newpage
\bibliographystyle{IEEEtran}
\typeout{}
\bibliography{references}

\end{document}